\documentclass{article}
\usepackage{spconf,amsmath,graphicx}
\usepackage{stfloats}
\usepackage{bbm}
\usepackage{spconf,amsmath,graphicx}
\usepackage{url}
\usepackage{amssymb}
\usepackage{amsmath,epsfig}
\usepackage{balance}
\usepackage{booktabs}
\usepackage{graphicx}
\usepackage{cite}
\usepackage{subfigure}
\usepackage{multirow}
\usepackage{hyperref}

\title{SKIP and SKIP: Segmenting Medical Images with Prompts}
\name{Jiawei Chen$^{1,2}$ \qquad Dingkang Yang$^{1,2}$ \qquad Yuxuan Lei$^{1,2}$ \qquad Lihua Zhang$^{1,2,3,*}$ 
\thanks{$^{*}$Corresponding author.}
\thanks{This work is supported by National Key R\&D Program of China (2021ZD0113503).}}

\address{
$^{1}$Academy for Engineering and Technology, Fudan University \\
$^{2}$Engineering Research Center of AI and Robotics, Ministry of Education, China \\
$^{3}$Jilin Provincial Key Laboratory of Intelligence Science and Engineering, Changchun, China \\
}
\begin{document}
%
\maketitle
\begin{abstract}
Most medical image lesion segmentation methods rely on hand-crafted accurate annotations of the original image for supervised learning. Recently, a series of weakly supervised or unsupervised methods have been proposed to reduce the dependence on pixel-level annotations. However, these methods are essentially based on pixel-level annotation, ignoring the image-level diagnostic results of the current massive medical images. In this paper, we propose a dual U-shaped two-stage framework that utilizes image-level labels to prompt the segmentation. In the first stage, we pre-train a classification network with image-level labels, which is used to obtain the hierarchical pyramid features and guide the learning of downstream branches. In the second stage, we feed the hierarchical features obtained from the classification branch into the downstream branch through short-skip and long-skip and get the lesion masks under the supervised learning of pixel-level labels. Experiments show that our framework achieves better results than networks simply using pixel-level annotations. 

\end{abstract}
\section{Introduction}
\label{sec:intro}

{M}{edical} image segmentation plays a vital role in clinical practice. Its purpose is to delineate the contour of organs, tissues, lesions, and other regions of interest from original medical images, such as CT, MRI, and pathological images. The segmentation of the lesion area is particularly important for image-guided radiotherapy. Benefiting from the development of deep learning technology, supervised learning methods based on convolutional neural networks (CNN) have shown good performance in a series of medical image segmentation tasks. However, CNN cannot capture the long-range dependencies between image pixels. With the advent of Transformers\cite{oktay2018attention}, researchers have migrated them to the vision domain and applied them to medical image segmentation\cite{chen2021transunet}. However, most of these methods\cite{lin2022ds}\cite{hatamizadeh2022unetr} still focus on using accurate pixel-level annotations, which require an extremely high medical level of annotators, and the annotation process is extremely cumbersome and expensive.

Therefore, a series of recent methods have been proposed to use less accurately labeled data for medical image lesion segmentation\cite{shen2023survey}. These methods can be summarized as follows: 1) Incomplete Supervision: Only part of the existing training data have per-pixel labels, and it can be subdivided into semi-supervision, partial supervision, and domain-specific supervision. 2) Coarse Supervision\cite{zhou2018brief}: All the training images are annotated, but the annotation of some images is coarse, and there are no pixel-by-pixel (PP) annotations. The labels can be divided into a) Image-Level: only category labels are provided for coarse-annotated images. b) Box-Level: Besides the class label, each image includes the object bounding box.

Currently, most research in this field focuses on incomplete supervision and centers on knowledge distillation\cite{tarvainen2017mean} methods, which learn through the combination of PP annotated data and the remaining unlabeled data. However, these works have strong limitations: as the proportion of unlabeled data increases, the performance of the model will degrade significantly. So when we want to improve the accuracy and generalization of the model, we need to increase the scale of labeled data and unlabeled data simultaneously. Additionally, these efforts have low utilization efficiency for samples without PP annotations and ignore the fact that most of the medical images clinically collected have useful knowledge, such as negative and positive discrimination conclusions. Considering most of the image data from the front line have already had preliminary clinical imaging diagnosis conclusions, how to utilize existing knowledge better should be especially worthy of our attention.

In this paper, we aim to establish an incomplete supervised learning model and propose a dual U-shaped two-stage framework named skip and skip (SKS). First, we train a branch in the framework using the previously ignored image-level coarse-grained labels in other work, enabling it to distinguish whether there is a lesion in the image and extract corresponding coarse-grained pyramid features. Subsequently, we perform fine-grained feature extraction based on fine-grained labels for lesion segmentation. Inspired by related modality interaction work~\cite{chen2024can, chen2024efficiency, chen2024miss, lei2023text}, we propose a novel module that combines the fine-grained annotation of medical images with the coarse-grained hidden knowledge, and the coarse-grained features are used to prompt lesion segmentation so that our framework acquires better results than the previous methods which only utilize PP annotations. The main contributions of this paper are as follows: 1) A dual U-shaped medical image lesion segmentation framework (SKS) is proposed. It utilizes the coarse-grained knowledge contained in vast amounts of medical images that have previously been overlooked. SKS achieves excellent segmentation results on small samples of data. 2) We propose a three-layer pyramid structure that employs Swin Transformer V2 (Swin-T v2) as its backbone to extract pyramid features of both coarse and fine grain from medical images within the same framework. 3) We implement three different skips that artfully integrate coarse- and fine-grained features within our framework. The skip of different features effectively prompts lesion segmentation results.

\begin{figure*}[ht]
    \centering
    \includegraphics[scale=0.43]{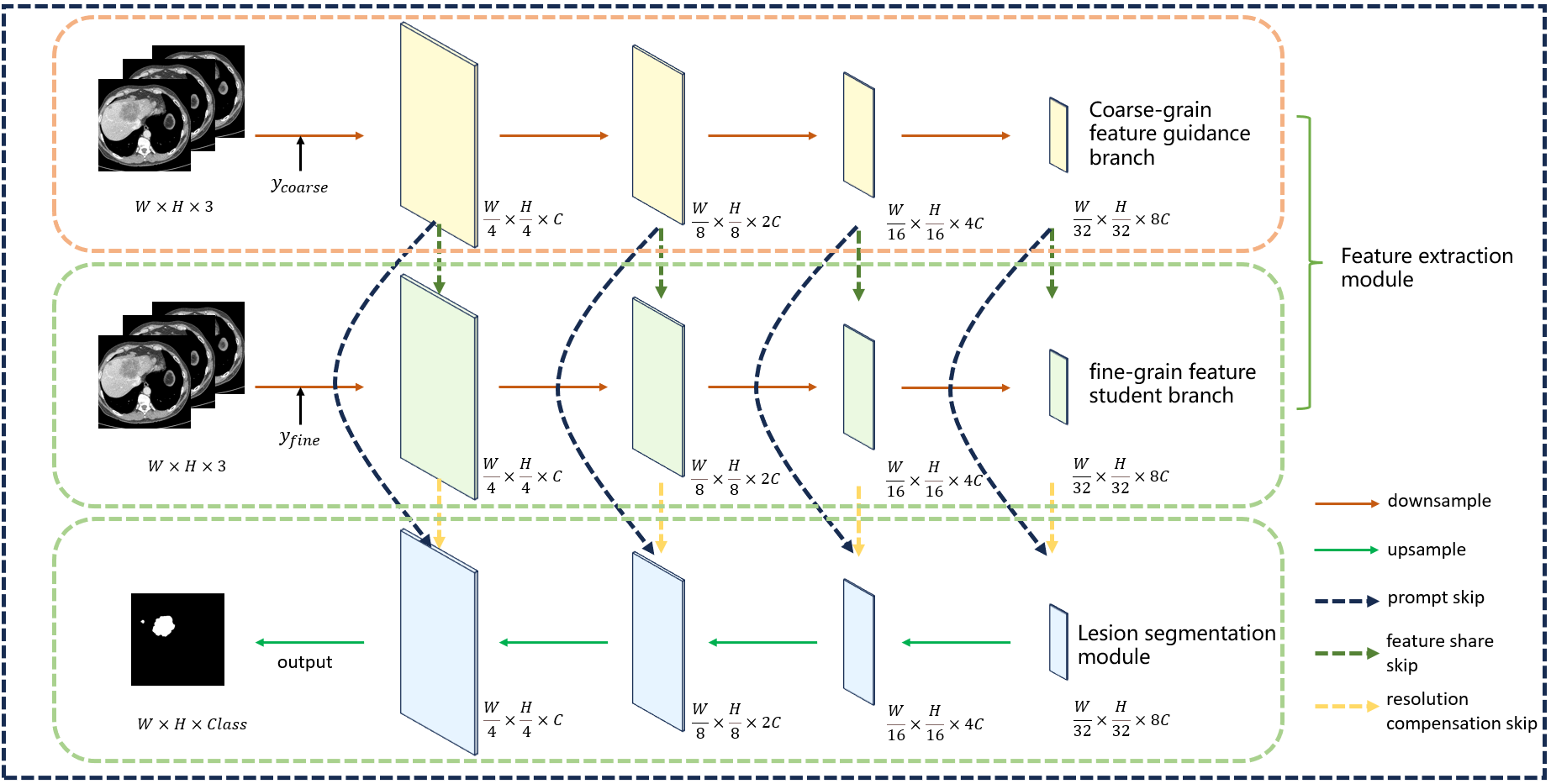}
    \caption{Overview of SKS. This dual U-shaped two-stage framework is a multi-level pyramid structure containing a feature extraction module and a lesion segmentation module, in the former represented by the yellow part is the coarse-grained feature guidance branch, and the green part is the fine-grained feature student branch.}
    \label{fig1}
    \vspace{-5pt}
\end{figure*}

\section{METHODOLOGY}
\label{2}

\subsection{SKS Overview}

To the best of our knowledge, SKS is the first framework using the existing coarse-grained diagnostic results of the medical image itself for lesion segmentation. Encouraged by the multi-task learning paradigm, we leverage the image features learned from the model's coarse-grained predictions to guide the downstream branches through three skips for lesion segmentation. As depicted in Figure \ref{fig1}, the model includes two main parts: a feature extraction module equipped with a coarse-grained image feature guidance branch and a fine-grained image feature student branch, and a lesion segmentation module. We assume that the input space is $\mathcal{X} = \{x_{1},x_{2},...,x_{n}\}$, $\mathcal{Y}_{cor} = \{y_{1}^{c},y_{2}^{c},...,y_{n}^{c}\}$ is the coarse-grained label (\textit{i.e.}, the clinical diagnosis result on the image level) while $\mathcal{Y}_{fine} = \{y_{1}^{f},y_{2}^{f},...,y_{n}^{f}\}$ is fine-grained label (\textit{i.e.}, the accurate annotation of the image lesion). Next, we will introduce the details of our SKS framework and how the upstream branch guides the segmentation module to perform lesion segmentation through three skips.

\subsection{Feature Extraction}
\label{2.2}
\begin{figure}[ht]
    \centering
    \includegraphics[scale=0.3]{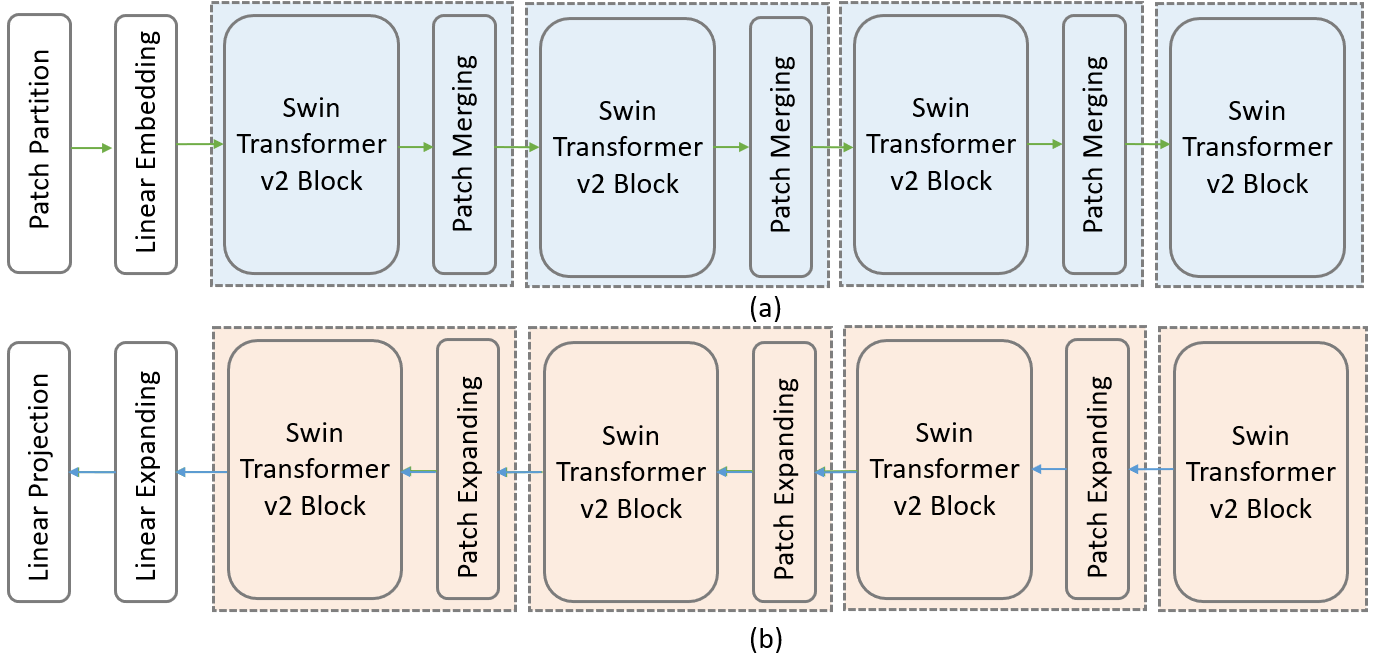}
    \caption{The structure of (a) the feature extraction module and (b) the lesion segmentation module.}
    \label{fig:enter-label}
    \vspace{-5pt}
\end{figure}

The feature extraction module includes two branches with similar structures: the coarse-grained image feature guidance branch and the fine-grained image feature student branch. To acquire prior knowledge from clinical image data itself, specifically the existing coarse-grained diagnosis results at the image level, we first feed the image into the coarse-grained image feature guidance branch. As shown in Figure~\ref{fig1}, the feature extraction modules are all pyramid structures. The length and width of the input image are respectively denoted as $H$ and $W$. We first divide the image into non-overlapping patches of size 4$\times$4 and each patch contains dimensions of 4$\times$4$\times$3 = 48 for its features. Details of these two branches are represented in Figure~\ref{fig:enter-label}(a). Following patch partitioning, a linear embedding layer maps the input to an arbitrary dimension $C$ before entering the pyramid feature extraction stage. Leveraging the advancements made by Swin-T v2, we employ a shifted window self-attention mechanism\cite{liu2021swin}: $Attention(Q, K, V ) = SoftMax(\frac{QK^{T}}{\sqrt{d}}+B)V$ to extract hierarchical features from the image, where $Q,K,V\in \mathbbm{R}^{M^{2}\times d}$ means query, key, and value matrices. $M^{2}$ is the number of patches in a window and $d$ is the dimension of query or key, and $B \in \mathbbm{R}^{M^{2}\times M^{2}}$ is a relative position bias. Within each pyramid feature extraction process, patch tokens are first processed through multiple Swin-T v2 blocks for representation learning. Then, a patch merging layer is utilized to downsample and $2\times$ increase sample dimensions while generating feature representations.
\subsection{Connection's Artful Leap} The SKS structure contains three skips, which we respectively refer to as: feature share skip, resolution compensation skip, and prompt skip. As shown in Figure \ref{fig1}, for the pyramid features from the $i$ layer $F_{pyra}^{i} = \{f_{1}^{i},f_{2}^{i},f_{3}^{i},f_{4}^{i}\}$, we fuse it with another pyramid feature $F_{pyra}^{k} = \{f_{1}^{k},f_{2}^{k},f_{3}^{k},f_{4}^{k}\}$, for arbitrary $f_{\lambda}^{fuse}$ in the new feature $F_{fuse}$, $f_{\lambda}^{fuse}=W\times(f_{\lambda}^{i}\oplus f_{\lambda}^{k})+b$, $W \in \mathbbm{R}^{C\times 2C}$ and $C$ represents the dimension of the $\lambda$ layer feature of the pyramid structure, $\oplus$ is concat and $b\in \mathbbm{R}^{C\times 1}$ is the bias.

\noindent \textbf{Feature Share Skip (FSS):} As mentioned in Section \ref{2.2} above, in the pre-training stage, we train the coarse-grained branch based on a large number of medical images with coarse-grained labels so that the branch could extract coarse-grained pyramid features of the image. In the training stage of the fine-grained branch, the coarse-grained pyramid feature is fused with this branch through the FSS, enabling it to learn new feature representations and combine coarse-grained features at the same time. As shown in Figure \ref{fig1}, when the fine-grained label of the image is used to train the fine-grained branch, the coarse-grained pyramid features are layered into each layer of the fine-grained branch through the FSS to guide this branch to simultaneously learn the coarse and fine-grained semantic information of the image. 

\noindent \textbf{Resolution Compensation Skip (RCS):} For the fused features, motivated by U-Net\cite{ronneberger2015u}, we send them into the lesion segmentation module by the RCS. This operation enables the lesion segmentation module to obtain the high-resolution information contained in the high-level feature map during the up-sampling process so as to obtain the details of the image better and improve the lesion segmentation accuracy. 

\noindent \textbf{Prompt Skip:}\label{2.3} The coarse-grained pyramid features are also fed into the lesion segmentation module through a prompt skip. Through this skip, we make the lesion segmentation module better able to analyze the information of the image based on the coarse-grained features of the image. Compared with previous works, this operation can greatly reduce the sample size of medical images with accurate annotations. The prompt and segmentation details will be described below. 

\subsection{Prompt and Segmentation}Here, the lesion segmentation of medical images is carried out under the prompt of the upstream branch. As introduced in Section \ref{2.3} above, for an image, the combined coarse- and fine-grained features obtained by the feature extraction module are sent to the lesion segmentation module through the RCS. The lesion segmentation module needs to up-sample the low-resolution feature map transmitted upstream into a mask. Details of it have been shown in Figure \ref{fig:enter-label}(b). Contrary to the patch merging layer of the feature extraction module, we use the patch expanding layer\cite{cao2022swin} to $2\times$ up-sample the deep features while reducing the feature dimension to generate the feature representation of the next layer. 

In addition to being connected to the fine-grained feature student module through the RCS, the lesion segmentation module also receives the coarse-grained pyramid features from the coarse-grained branch through the prompt skip. After the supervised learning of the previous coarse-grained label, the coarse-grained branch can provide the lesion segmentation module with a prior coarse-grained feature representation. With the prompts of this feature representation, the lesion segmentation module directly obtains certain feature representations that previous segmentation models required learning through some samples. At the same time, the task branch of the model can focus on the segmentation of the lesions without needing to distinguish whether a medical image contains lesions, reducing the sample size of accurate annotations required to achieve the same segmentation effect.

\section{EXPERIMENTS}
\subsection{Experiment Setup}
\textbf{Dataset Configuration:} Considering the complexity and diversity of lesions at the liver site, we choose to evaluate our framework on the LITS dataset\cite{DBLP:journals/corr/abs-1901-04056}, which contains a total of 200 CT scans from various clinical sites. Each CT scan contains about 100-400 CT slices, and every piece of CT slices has been experienced clinician's manual annotation of liver and liver tumour location. In this paper, we set the window height and width as [-200, 200] to extract the liver region. For each slice in the CT scan, we take the upper and lower adjacent slices of it as a single input $x_{i}$ and the 3D input image size is 224 $\times$ 224. We consider whether a slice has a lesion as its coarse-grained label and the precise annotation of the slice as its fine-grained label.

\noindent \textbf{Training Details:}
For evaluation of the LITS dataset, firstly, we train the coarse-grained branch with 131 CT scans. As described in Section \ref{2}, we employ 3D slices and coarse-grained labels to carry out supervised training in this branch, enabling it to identify whether the abdominal CT has liver lesions and extract the coarse-grained pyramid features. Since SKS aims to achieve the same segmentation effect with fewer samples, we randomly selected 35 CT scans from the original dataset to train the fine-grained branch and lesion segmentation branch for segmentation. We train all models with dice loss\cite{DBLP:journals/corr/MilletariNA16} and validate the effect on 8 CT scans. For the Swin-T v2 backbone, the patch size is 4 $\times$ 4 and every layer in the pyramid structure has two Swin-T v2 blocks. 

\begin{figure}[htbp]
    \centering
    \includegraphics[scale=0.44]{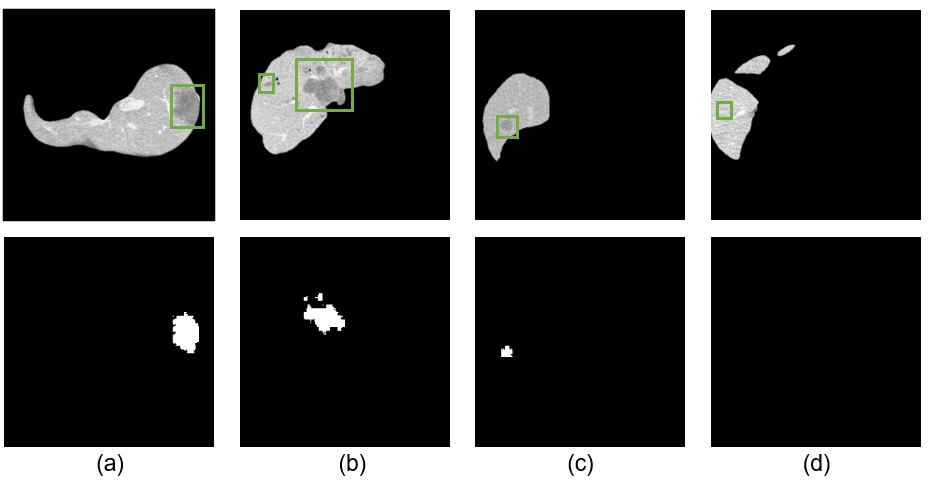}
    \caption{Visualization of liver segmentation results. The images above are the 2D slices of the liver region with the lesions outlined in green boxes, and the images below show the segmentation results of our framework. }
    \label{fig3}
    \vspace{-8pt}
\end{figure}
\subsection{Experiment  Evaluation}
In this section, we show that SKS can effectively leverage coarse-grained knowledge from images and achieve excellent performance with a small number of samples compared to previous supervised training methods. As shown in Table \ref{tab:mytable}, with an upstream coarse-grained branch achieving an accuracy of 89\% in diagnosing liver lesions, the SKS framework achieved a global DSC (Dice Similarity Coefficient) of 0.549 for liver tumor segmentation using only 35 CT scans with precise annotations. In contrast, U-Net and U-Net++\cite{zhou2019unet++} achieve DSC of only 0.489 and 0.509 using the same 35 CT scans. Therefore, compared to traditional segmentation models, our SKS framework significantly achieves better results with a small amount of samples. The segmentation masks are visualized in Figure \ref{fig3}, which shows SKS have demonstrated good segmentation performance for lesions of different sizes. We extract the liver region from the CT image and delete other unrelated organ regions to make the segmentation result look clearer. In Figure \ref{fig3}(d), the tumor is too small and similar to the background, so SKS is unable to recognize it.

\begin{table}[t]
\setlength{\tabcolsep}{4pt}
\centering
\caption{Comparison and ablation experiments on the LITS dataset. ``JC'' means the Jaccard Coefficient. ``w/o'' means the without.}
\resizebox{\linewidth}{!}{%
\begin{tabular}{l|ccccc}
\hline
Method                   & DSC   & JC & Precision & Recall \\ \hline
U-Net \cite{ronneberger2015u}                        & 0.489 & 0.379               & 0.678     & 0.451  \\
U-Net++ \cite{zhou2019unet++}                       & 0.509 & 0.379               & 0.643     & 0.524  \\
Att-UNet \cite{DBLP:journals/corr/abs-1804-03999}                     & 0.348 & 0.246               & \textbf{0.727}     & 0.258  \\
CE-Net \cite{gu2019net}      & 0.401& 0.286 &0.354 &0.592 \\
CE-Net-OCT \cite{gu2019net}      & 0.468& 0.342 &0.490 &0.527 \\
Channel-UNet \cite{2019Channel}                     & 0.524 & 0.392               & 0.508     & 0.591  \\
SwinUnet \cite{cao2022swin}                     & 0.475 &0.355                     &0.365           &\textbf{0.952}        \\ \hline
w/o coarse-grain branch       & 0.420      & 0.311          &0.686      & 0.347   \\
w/o prompt skip               & 0.505      & 0.379          & 0.728     & 0.447       \\
w/o RCS                      &  0.467     &  0.331                   & 0.400          &  0.672      \\ \hline
SKS (ours)                    & \textbf{0.549} & \textbf{0.456 }              & 0.542     & 0.580  \\ \hline
\end{tabular}%
}
\label{tab:mytable}
\vspace{-5pt}
\end{table}

\subsection{Ablation Study}

\label{sec:pagestyle}

We conduct ablation experiments on the key modules of the proposed SKS framework, and the results are presented in the lower part of Table \ref{tab:mytable}. To evaluate the impacts of the FSS, the entire coarse-grained branch is removed and the remaining parts of the framework cannot recognize lesions due to the lowest recall of 0.347. When only the FSS and RCS are present, our framework achieves a DSC of 0.505. Moreover, removing the RCS results in a decrease of DSC to 0.467. The ablation studies demonstrate that the use of our coarse-grained feature guidance branch and SKS structure can fully extract coarse-grained knowledge from medical images, resulting in a better understanding of the images and achieving a better segmentation effect with few samples compared with other methods.

\section{CONCLUSION}
\label{sec:typestyle}
In this paper, we propose a dual U-shaped two-stage framework for medical image segmentation. This framework extracts both fine-grained and coarse-grained knowledge from medical images through a multi-level pyramid structure and introduces a Connection's Artful Leap to fuse the coarse-grained and fine-grained features, utilizing the coarse-grained knowledge to guide the downstream branch for lesion segmentation. Experiments show that SKS realizes a better segmentation effect with few samples. In the future, we will attempt to extend the SKS framework to more signal fields.



\small

\end{document}